\documentclass{article} 
\usepackage{iclr2027_conference,times}

\usepackage{amsmath,amsfonts,bm}

\def\eqref#1{equation~\ref{#1}}

\def\1{\bm{1}}

\DeclareMathAlphabet{\mathsfit}{\encodingdefault}{\sfdefault}{m}{sl}
\SetMathAlphabet{\mathsfit}{bold}{\encodingdefault}{\sfdefault}{bx}{n}

\usepackage{booktabs}
\usepackage{hyperref}
\usepackage{url}
\usepackage{multirow}
\usepackage{algorithm}
\usepackage{algpseudocode}
\usepackage{amsmath}
\usepackage{bm}
\usepackage{graphicx}
\usepackage{wrapfig}

\title{Autonomous Fashion Outfit Composition via Unified Aesthetic Foresight Model}

\author{%
  Kaicheng PANG\\
  Laboratory for Artificial Intelligence in Design\\
  Hong Kong SAR\\
  \texttt{kcpang@aidlab.hk} \\
  \And
  Xingxing ZOU \\
  Hong Kong Polytechnic University\\
  Hong Kong SAR \\
  \texttt{xingxing.zou@polyu.edu.hk} \\
  \AND
  Ruohan XU \\
  The University of Queensland \\
  Brisbane, Australia \\
  \texttt{ruohan.xu@student.uq.edu.au} \\
  \And
  Waikeung Wong\thanks{Corresponding author.} \\
  Laboratory for Artificial Intelligence in Design \\
  Hong Kong SAR \\
  \texttt{calvinwong@aidlab.hk} \\
}

\iclrfinalcopy 
\begin{document}

\maketitle

\begin{figure*}[h]
    \centering
    \includegraphics[width=\linewidth]{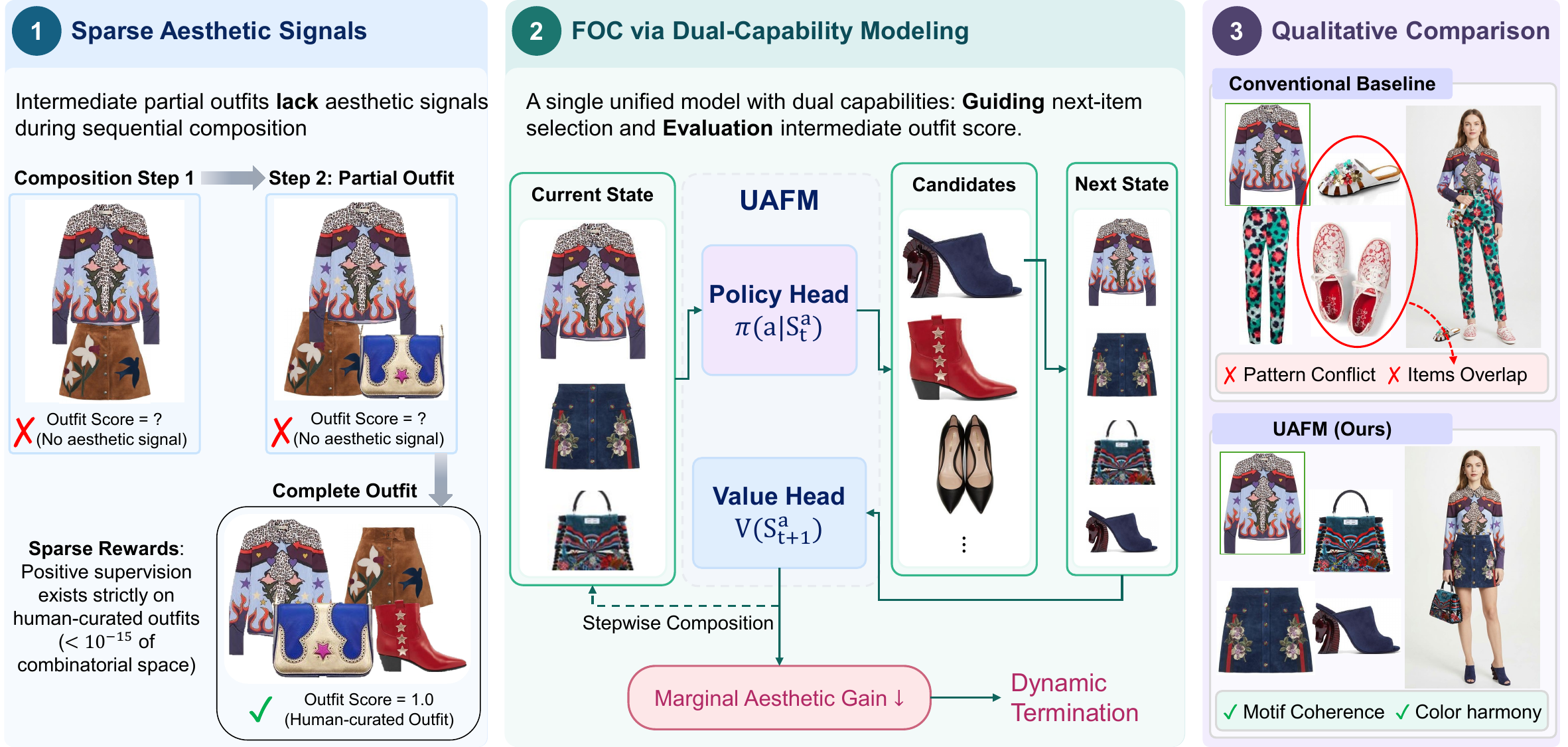}
    \vspace{-5mm}
    \caption{\textbf{Fashion Outfit Composition: Challenge and Paradigm.} 
    Conventional paradigms face severe reward sparsity during sequential composition, as positive supervision is restricted to complete human-curated outfits, leaving intermediate partial states unmonitored (Panel 1). 
    Addressing FOC requires dual capabilities: bootstrapping intermediate state values from sparse terminal rewards and predicting next-item intent. Unifying intermediate aesthetic evaluation (Value Head) and candidate selection (Policy Head) onto a shared backbone enables autonomous dynamic termination by tracking diminishing marginal aesthetic gains (Panel 2). 
    Our method generates cohesive ensembles, whereas conventional baselines suffer from severe pattern conflicts and item overlaps (Panel 3).}
    \label{fig:teaser}
\end{figure*}

\begin{abstract}
Fashion Outfit Composition (FOC) requires sequentially assembling fashion items into a stylistically cohesive ensemble. Existing works struggle to model this step-by-step process effectively, primarily because they fail to jointly optimize the two critical capabilities required for FOC: intermediate outfit value evaluation and complementary item prediction. This structural disconnect, compounded by the severe sparsity of step-wise aesthetic signals, leaves them without a mechanism to autonomously determine when to stop the composition process. To address these challenges, we introduce the Unified Aesthetic Foresight Model (UAFM), which seamlessly unifies both capabilities into a dual-head architecture over a shared backbone. Crucially, we formulate FOC as a deterministic Markov Decision Process and optimize a value head via a post-decision state temporal difference (TD) objective to recursively backpropagate sparse terminal aesthetic rewards to intermediate states. This enables the model to accurately estimate the potential of a partial outfit evolving into a compatible ensemble. Consequently, UAFM can compute step-wise marginal aesthetic gains for dynamic termination. Furthermore, this unified design strictly aligns aesthetic evaluation and complementary item prediction within a shared representation manifold, intrinsically regularizing the combinatorial search space. Extensive experiments on the Polyvore-Outfits dataset demonstrate that UAFM establishes a new state-of-the-art across both FOC and conventional fashion tasks. Extensive ablation studies further confirm that our post-decision state TD formulation provides the aesthetic value prediction necessary for autonomous dynamic termination, while validating the synergistic benefits of our unified dual-head architecture.
\end{abstract}

\section{Introduction}
\label{sec:intro}
Fashion Outfit Composition (FOC) plays an increasingly pivotal role in modern retail ecosystems, directly driving units per transaction by encouraging cross-category basket completion~\citep{mameli2021deep}. Unlike standard point-wise item recommendation~\citep{song2023mm,zhu2024collaborative}, FOC requires \textbf{sequentially} assembling an \textbf{unordered set} of items from a massive garment library into a stylistically cohesive ensemble. This process needs to balance structural constraints, such as avoiding redundant categories, with visual harmony grounded in collective human aesthetics.
Prior works foundational to outfit composition mainly focused on modeling overall outfit compatibility as a static classification task, utilizing subspace projections~\citep{vasileva2018learning}, graph neural networks~\citep{cui2019dressing,li2020hierarchical}, and hierarchical relation networks~\citep{pang2025towards}. This field later evolved to incorporate graph attention networks~\citep{saed2025hybrid} and large multimodal models~\citep{chang2025using}. However, these methods act as passive evaluators, incapable of actively retrieving complementary items from vast candidate pools. 
More recently, retrieval-based frameworks have been proposed to tackle complementary item selection by predicting continuous latent embeddings~\citep{tan2019learning, sarkar2023outfittransformer}. Another branch of generative methods~\citep{xu2024diffusion, yu2025fashiondpo} focuses on synthesizing pixel-level outfit images constrained by explicit reference masks, category prompts or utilizing image-to-image translation for retrieval templates~\citep{attimonelli2024fashion}. However, such one-shot approaches create a severe gap between unconstrained visual concepts and real-world inventory, frequently resulting in retrieved physical items that are mutually incompatible.

Instead, FOC is inherently a sequential decision-making process that hinges on \textbf{aesthetic foresight}—a step-wise dual capability to evaluate the aesthetic potential of partial states and select complementary candidates from the library.
When existing methods are tasked with this sequential composition, they reveal three critical bottlenecks. 
First, accurately evaluating the aesthetic potential of a partial outfit is inherently plagued by \textbf{sparse rewards} as illustrated in Figure~\ref{fig:teaser} (Panel 1). Positive signals only exist for \textbf{human-curated complete} outfits, which constitute a vanishingly small fraction of the combinatorial space. Traditional classification-based scoring paradigms~\citep{vasileva2018learning}, trained solely on fully assembled outfits, suffer from a severe train-test mismatch and fail to assess whether a partial outfit can eventually yield a compatible ensemble. 
Second, existing frameworks~\citep{sarkar2023outfittransformer} typically treat the demanding dual capabilities as disparate models optimized via decoupled objectives. This architectural isolation induces conflicting latent spaces, leading to compounding errors along the composition trajectory.
Third, because autonomous composition operates without predefined outfit schemas, current approaches lack a formalized stopping mechanism, leading to unbounded over-composition as shown in Figure~\ref{fig:teaser} (Panel 3).

To overcome these gaps, we propose the \textbf{Unified Aesthetic Foresight Model (UAFM)} (schematically illustrated in Figure~\ref{fig:teaser}, Panel 2), which unifies the dual capabilities onto a single shared backbone by equipping it with a \textbf{policy head} for complementary item retrieval and a \textbf{value head} for aesthetic potential evaluation. 
The core innovation of UAFM lies in shifting fashion compatibility modeling from static binary classification to state-value function estimation, explicitly modeling the expected cumulative aesthetic return of partial outfits within a Markov Decision Process (MDP).
Specifically, we reformulate compatibility learning through a novel post-decision state temporal difference (TD) objective to train its value head. By defining the post-decision state as the deterministic ensemble immediately following an item addition, UAFM decouples candidate evaluation from environmental transitions, enabling sparse terminal aesthetic rewards to recursively backpropagate to early intermediate steps. This structural credit assignment endows the model with \textit{aesthetic foresight} to evaluate the long-term potential of incomplete outfits.
To complement value estimation, we optimize the policy head via contrastive learning. This explicitly aligns the predicted intent embedding with the ground-truth item selected by human within the shared latent space. 
During inference, at each composition step, we leverage UAFM's policy head to retrieve a top candidate subset from the library and employ the value head to evaluate the aesthetic potential of each augmented post-decision state within a beam search framework. As the ensemble expands and approaches aesthetic completion, the composition naturally halts when the computed marginal aesthetic gain falls below a predefined threshold, achieving autonomous FOC without explicit outfit schemas.

Extensive experiments on the Polyvore-Outfits dataset~\citep{vasileva2018learning} validate UAFM. In FOC, UAFM demonstrates strong aesthetic quality and structural consistency compared to existing baselines, while maintaining low inference latency. Additionally, UAFM shows consistent improvements over prior methods on two conventional discriminative tasks: fashion compatibility evaluation and complementary item retrieval. Comprehensive ablation studies confirm that our post-decision state TD objective enables effective dynamic termination, and that the unified architecture yields consistent performance gains.
The main contributions of this work are summarized as follows:
\begin{itemize}
    \item To the best of our knowledge, we are the first to formalize Fashion Outfit Composition as a deterministic MDP, proposing a post-decision state TD objective to recursively backpropagate sparse aesthetic rewards for evaluating partial outfits.
    \item We propose the Unified Aesthetic Foresight Model (UAFM), which seamlessly synergizes aesthetic potential evaluation and complementary item prediction within a single shared-backbone, dual-head framework.
    \item By leveraging step-wise marginal aesthetic gains to dynamically terminate composition, UAFM achieves autonomous outfit composition while establishing new state-of-the-art results across both FOC and discriminative benchmarks.
\end{itemize}

\section{Methodology}
\subsection{Problem Formulation} 
\label{subsec:problem_formulation}
We define the task of Fashion Outfit Composition (FOC) as iteratively selecting garments from a large-scale library $\mathcal{X}$ to construct a stylistically cohesive, unordered set $S \subseteq \mathcal{X}$, equipped with the capability to autonomously halt upon completeness. This sequential, self-terminating composition naturally maps to a fully-observed, deterministic Markov Decision Process (MDP), wherein we formulate the state at step $t$ as the currently assembled outfit, $S_t = \{a_0, a_1, \dots, a_{t-1}\}$. The state space is thus defined over the power set of the library: $\mathcal{S} \subseteq \mathcal{P}(\mathcal{X}) \setminus \{ \emptyset \}$. Owing to deterministic environmental transitions, taking an action $a_{t} \in \mathcal{X} \setminus S_t$ uniquely specifies the next state, which we formally represent via the \textit{post-decision state}~\citep{sutton1998reinforcement}, $S_t^a \doteq S_t \cup \{a_t\}$. Crucially, this post-decision formulation collapses the high-dimensional action-value function $Q(S_t, a_t)$ over a massive library into a simplified state-value function $V(S_t^a)$ directly over the augmented outfit. The composition trajectory is guided by a policy $\pi(a_{t+1} \mid S^a_t)$ that actively retrieves complementary candidates $a_{t+1} \in \mathcal{X}$ to incrementally expand the ensemble.

\begin{figure}[t]
    \centering
    \includegraphics[width=1.0\linewidth]{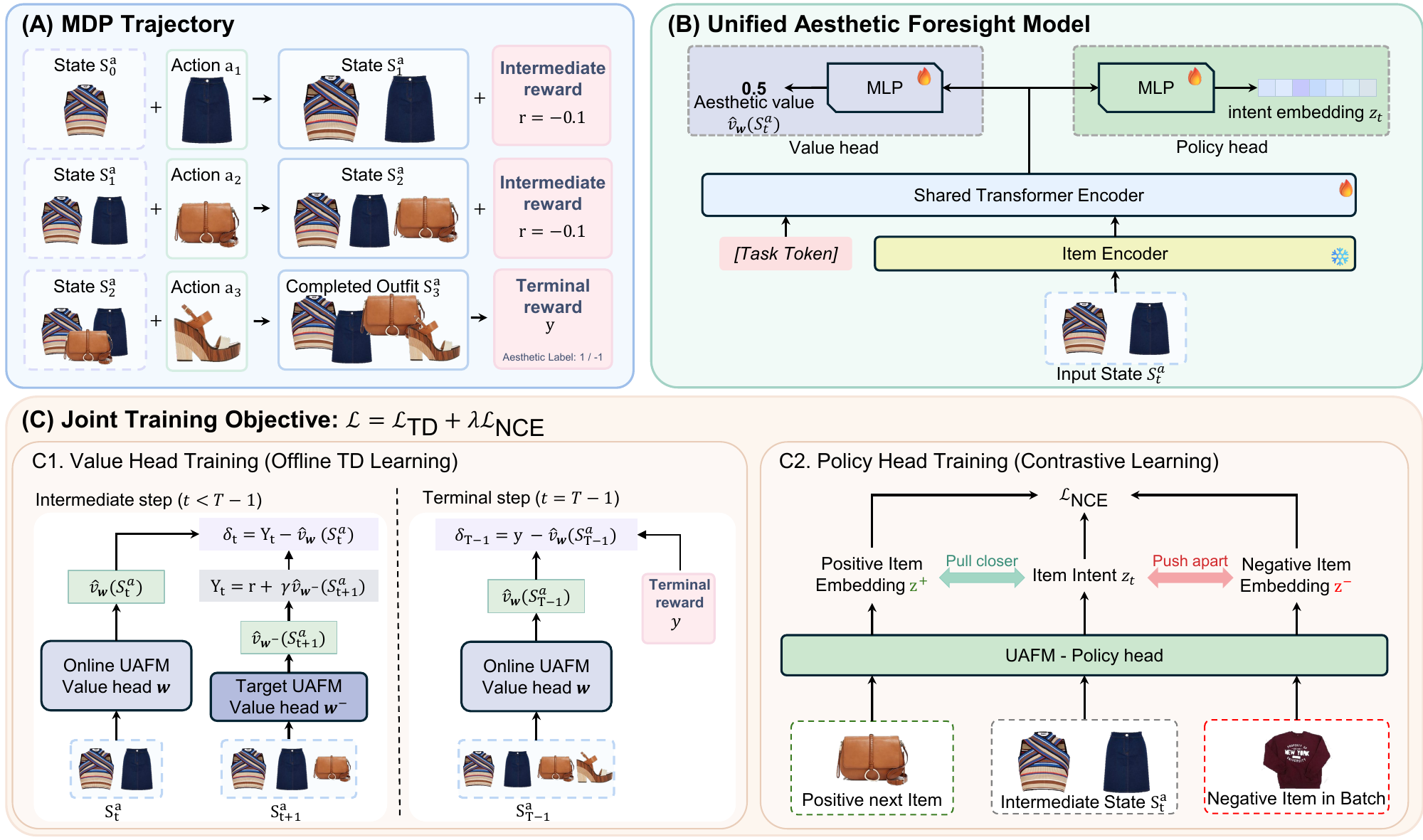}
    \vspace{-5mm}
    \caption{\textbf{Overview of the UAFM architecture and training pipeline.}
    \textbf{(A) MDP Trajectory:} Human-curated outfits are decomposed into state-action trajectories, governed by step penalties and terminal aesthetic rewards. 
    \textbf{(B) UAFM Architecture:} A shared encoder backbone jointly parameterizes the value head (aesthetic potential) and the policy head (stylistic intent) within a dual-head framework. 
    \textbf{(C) Joint Training Objective:}
    \textit{(C1) Value Head Training} employs a proposed post-decision state TD objective to recursively backpropagate sparse terminal rewards to early intermediate states; \textit{(C2) Policy Head Training} aligns latent intents via InfoNCE-based contrastive learning.}
    \label{fig:model_structure}
\end{figure}

To explicitly operationalize this MDP framework, we decompose outfits into sequential decision trajectories of state-action transitions, as illustrated in Figure~\ref{fig:model_structure} (Part A). 
Starting from an initial state $S_0^a$, the environment's inherent feedback is strictly sparse. To govern the step-wise assembly, we impose a constant \textbf{step penalty} of $r = -0.1$ on each intermediate transition. A definitive global aesthetic reward $y \in \{+1.0, -1.0\}$ is emitted exclusively upon reaching the completed outfit state ($S_3^a$ in this case). Positive terminal outfits are sourced directly from the dataset, while negative ones are constructed as described in Appendix~\ref{apx:appendix_noc}. During autonomous composition, the MDP sequence dynamically terminates upon reaching aesthetic saturation where the marginal aesthetic gain of adding an extra item is outweighed by its compositional cost.

\subsection{Architecture of Unified Aesthetic Foresight Model}
As depicted in Part B of Figure~\ref{fig:model_structure}, UAFM is designed to process an unordered set of garments, inherently preserving permutation invariance. Specifically, the constituent items of an input state $S_t^a$ are first projected into continuous feature representations via a frozen pre-trained item encoder. To enable the network to execute distinct objectives given identical input, we prepend learnable, task-specific tokens (\textit{e.g.}, \texttt{<VAL\_TOKEN>} for value head and \texttt{<POL\_TOKEN>} for policy head) to these concatenated item embeddings.
These embeddings are subsequently processed by a shared Transformer~\citep{vaswani2017attention} encoder backbone, which inherently guarantees permutation invariance over the item set by operating without positional encodings.
This unified architecture not only reduces parameter redundancy but also promotes multi-task synergy by cultivating a joint representation manifold. 
The contextualized embedding of the prepended task token is extracted from the final hidden layer and fed directly into the respective value or policy head.
The specific training pipelines for UAFM are elaborated in the following subsections.

\subsubsection{Aesthetic Value Prediction via Post-Decision State TD Objective}
\label{subsec:TD_learning}
The value head $\hat{v}_\mathbf{w}(\cdot) : \mathcal{S} \rightarrow \mathbb{R}$, parameterized by $\mathbf{w}$, estimates the expected aesthetic value of a given post-decision state. Specifically, for state-action trajectories in the empirical distribution $\mathcal{D}$ terminating at either positive or negative outfits, the target value $v(S_t^a)$ is recursively defined via the Bellman expectation equation:
\begin{equation}
     \hat{v}_\mathbf{w}(S_t^a) \approx v(S_t^a) = \mathbb{E}_{\mathcal{D}} \left[ R_{t+1} + \gamma v(S_{t+1}^a) \mid S_t^a \right],
\end{equation}

where $\gamma \in (0, 1]$ is the discount factor. An immediate step penalty of $R_{t+1} = -0.1$ is explicitly specified for intermediate transitions ($S_t^a \to S_{t+1}^a$ for $t < T-1$).
To resolve the mathematical contradiction between intermediate step predictions and terminal aesthetic evaluations, we introduce an absorbing terminal state $S_{T}$ with $v(S_{T}) \equiv 0$.
At the final step $t = T-1$, the completed outfit $S_{T-1}^a$ naturally transitions to $S_{T}$, triggering the environment to emit the ground-truth aesthetic label $R_T = y$ as the terminal reward. Thus, the expected value of the terminal post-decision state aligns with the given label: $v(S_{T-1}^a) = R_T + \gamma v(S_{T}) = y$.

Under this formulation, the Bellman expectation equation directly translates to the following empirical TD target $Y_t$ as illustrated in Figure~\ref{fig:model_structure}-C1:
\begin{equation}
Y_t =  
\begin{cases}  
y, & \text{if } t = T - 1 \quad \text{(Terminal transition to } S_{T}\text{)} \\   
r + \gamma \hat{v}_{\mathbf{w}^-}(S_{t+1}^a), & \text{if } t < T - 1 \quad \text{(Intermediate transition)}  
\end{cases}
\end{equation}
Here, we employ a target network $\mathbf{w}^-$, periodically updated from the online network $\mathbf{w}$, to stabilize offline value function learning~\citep{volodymyr2015human}.
The value network is optimized by minimizing the squared TD error 
\begin{equation}
    \mathcal{L}_{\text{TD}}(\mathbf{w}) = \frac{1}{2} \mathbb{E}_{(S_t^a, S_{t+1}^a, R_{t+1}) \sim \mathcal{D}} \left[ \left(\hat{v}(S_t^a; \mathbf{w}) - Y_t\right)^2 \right],
\end{equation}
where $\mathcal{D}$ is the trajectory distribution of the training data~\citep{kumar2020conservative}. This loss seamlessly unifies two complementary supervision signals:
(1) For the terminal state ($t = T-1$), the TD error is defined as $\delta_{T-1} = y - \hat{v}(S_{T-1}^a; \mathbf{w})$. This term quantifies the discrepancy between the predicted harmony and the ground-truth label, effectively acting as a supervised aesthetic regressor for complete outfits.
(2) For intermediate states ($t < T-1$), the TD error is $\delta_t = \left( r + \gamma \hat{v}(S_{t+1}^a; \mathbf{w}^-) \right) - \hat{v}(S_t^a; \mathbf{w})$. Through this recursive mechanism, the explicit aesthetic reward anchored at the terminal state is systematically propagated backward along the trajectory: $y \to \hat{v}(S^a_{T-1}) \to \dots \to \hat{v}(S^a_1)$.

\subsubsection{Complementary Item Prediction via Contrastive Learning} 
The objective of the policy head is to capture fine-grained complementary relationships between existing selections and target items. Specifically, given a partial outfit $S^a_t$, the policy head embeds this post-decision state into a latent representation $\mathbf{z}_t \in \mathbb{R}^d$. This query represents the target stylistic intent, serving as a continuous-space prior for the next selection. Conversely, each candidate item $x_i \in \mathcal{X}$ in the library is projected into the same latent manifold by passing its visual feature through the shared backbone and policy head \textit{without} prepending the task token, obtaining an indexing embedding $\mathbf{z}_i \in \mathbb{R}^d$. This unified design ensures that the predicted intent $\mathbf{z}_t$ and candidate item embeddings $\mathbf{z}_i$ are strictly aligned in a shared semantic space, inducing a policy $\pi(a \mid S_t^a)$ via efficient nearest-neighbor retrieval.
We optimize this alignment via contrastive learning using the InfoNCE loss ($\mathcal{L}_{\text{NCE}}$), as illustrated in Figure~\ref{fig:model_structure}-C2. During training, given an intermediate state $S^a_t$, the policy head extracts the predicted item intent embedding $\mathbf{z}_t$. The contrastive objective explicitly pulls $\mathbf{z}_t$ closer to the positive item embedding $\mathbf{z}^+$ (derived from the human selection) while pushing it apart from in-batch negative item embeddings $\mathbf{z}^-$. This contrastive pulling-and-pushing mechanism forces the predicted intent vector to reside in close proximity to stylistically compatible candidates within the shared latent manifold.
We jointly optimize UAFM for both tasks, and the total end-to-end training objective is: $\mathcal{L} = \mathcal{L}_{\text{TD}} + \lambda \mathcal{L}_{\text{NCE}}$, where $\lambda$ is a trade-off hyperparameter.

\subsection{End-to-End Outfit Composition Pipeline}
\label{subsec:generation_pipeline} 
By leveraging UAFM for both value estimation and item prediction, the composition process can be executed autonomously. Starting from an initial state $S^a_0$, at each step $t$, the policy head first derives a composition intent vector $\mathbf{z}_t$ to retrieve a top-$K$ candidate subset from $\mathcal{X}$, denoted as $\mathcal{K}_t \subset \mathcal{X}$. 
Subsequently, the value head evaluates the augmented post-decision states $S^a_t \cup \{a\}$ for all retrieved candidates $a \in \mathcal{K}_t$.  We employ a standard beam search algorithm to explore high-potential trajectories. The composition process halts naturally if the maximum predicted value improvement of the next augmented state falls below a predefined aesthetic threshold: $\max_{a\in \mathcal{K}_t} (\hat{v}(S^a_t \cup \{a\}) - \hat{v}(S^a_t) ) < \tau_\text{stop}$, signaling that the ensemble has achieved aesthetic saturation. By dynamically identifying when an outfit reaches aesthetic completeness, this mechanism enables autonomous FOC across variable outfit lengths without hardcoded rules.

\section{Experiments}
\label{sec:experiments}
We comprehensively evaluate the capabilities and effectiveness of UAFM across three key dimensions:
(1) fashion outfit composition quality (\S\ref{subsec:generation_quality}),
(2) downstream discriminative performance in fashion compatibility prediction and complementary item retrieval (\S\ref{subsec:sota_comparison}), and
(3) ablation studies validating the essential roles of our post-decision state TD objective and unified architecture (\S\ref{sec: ablation})

\subsection{Evaluation on Fashion Outfit Composition Task}
\label{subsec:generation_quality}
\textbf{Experimental Setup:} Since no existing work is specifically tailored to directly address FOC, we adapt strong baselines from related fashion tasks for comparison. To thoroughly evaluate UAFM, we benchmark it against these adapted methods on the Polyvore-Outfits dataset~\citep{vasileva2018learning}.
The composition task is initialized with a seed item sampled from a human-curated reference outfit in the test set (denoted as GT outfit for brevity). Our test benchmark comprises $9,000$ evaluation cases with GT outfit lengths ranging from $3$ to $6$, where initial seed items are selected strictly from major categories (\textit{i.e.}, \textit{tops}, \textit{bottoms}, \textit{all-body}, \textit{outerwear}, and \textit{shoes}). To ensure a fair comparison, all evaluated baselines are provided with the exact ground-truth outfit length as prior knowledge, thereby compensating for existing baselines' lack of dynamic termination capabilities.

\textbf{Metrics.} The assembled outfits are evaluated across the following metrics:
(1) \textbf{Human Preference (H. Pref.):} Win rate in a blind A/B human study (detailed in Appendix~\ref{apx:human_study}).
(2) \textbf{VLLM Aesthetic Score (VAS):} Adopting the LLM-as-a-Judge paradigm~\citep{zheng2023judging}, the average score on a $1$--$5$ scale evaluated by Gemini-3.5-flash across five aesthetic dimensions, with GT outfits achieving a baseline score of $3.55$.
(3) \textbf{Category Alignment (Cate. Align):} Measures the structural category consistency between composed and GT outfits. Higher rates reflect stronger structural fidelity to human-curated rules.
(4) \textbf{Category Duplication Rate (Cate. Dup.):} Quantifies categorical duplication within composed outfits. A lower percentage indicates superior structural diversity and adherence to realistic fashion composition rules, where the GT outfits reach the low rate at $2.01\%$.
(5) \textbf{Visual Redundancy Rate (Vis. Red.):} Measures fine-grained visual item-level redundancy using normalized CLIP image embeddings. We compute the fraction of item pairs whose cosine similarity exceeds a predefined threshold of $0.75$. Lower scores demonstrate that the model retrieves visually distinct items rather than near-identical duplicates, with GT outfits achieving a minimal baseline rate of $0.15\%$.
(6) \textbf{Latency:} Measures the average composition time per outfit in seconds.

\begin{table}[t]
\centering
\caption{Results of composed outfits quality across different baselines. \textbf{Bold} and \underline{underline} indicate the best and second-best performances, respectively. Metrics with $\downarrow$ denote lower values are better. Scores for \textit{VAS}, \textit{Cate. Align (\%)}, \textit{Cate. Dup. (\%)}, and \textit{Vis. Red. (\%)} report 5-fold Mean $\pm$ SE.}
\label{tab:generation_quality}
\setlength{\tabcolsep}{5pt}
\begin{tabular}{lcccccc}
\toprule
Method & H. Pref. & VAS  & Cate. Align & Cate. Dup. $\downarrow$ & Vis. Red. $\downarrow$ &  Latency (s) $\downarrow$\\
\midrule
Random & - & 2.50 ± 0.04 & 42.34 ± 0.15 & 17.84 ± 0.13 & \textbf{0.18 ± 0.03} & \textbf{0.0078} \\
Type-Aware & - & 2.69 ± 0.04 & 41.15 ± 0.28 & 21.46 ± 0.11 & 1.39 ± 0.08 & 0.15   \\
OutfitTrans. & 25.75\%  & 3.18 ± 0.03 & 62.21 ± 0.23 & 20.58 ± 0.25 & 6.10 ± 0.11 & 0.14   \\
Gemini-3.5 & \underline{28.08\%}   & \underline{3.28 ± 0.03} & \underline{64.93 ± 0.23} & \underline{15.48 ± 0.17} & 8.60 ± 0.15 & 6.35   \\
\midrule
\textbf{UAFM} & \textbf{46.17\%} & \textbf{3.57 ± 0.02} & \textbf{72.18 ± 0.27} & \textbf{5.69 ± 0.15  } & \underline{1.20 ± 0.09 }  & \underline{0.13}\\
\bottomrule
\end{tabular}
\vspace{-7mm}
\end{table}

\begin{wrapfigure}{r}{0.45\textwidth}
  \centering
  \includegraphics[width=0.45\textwidth]{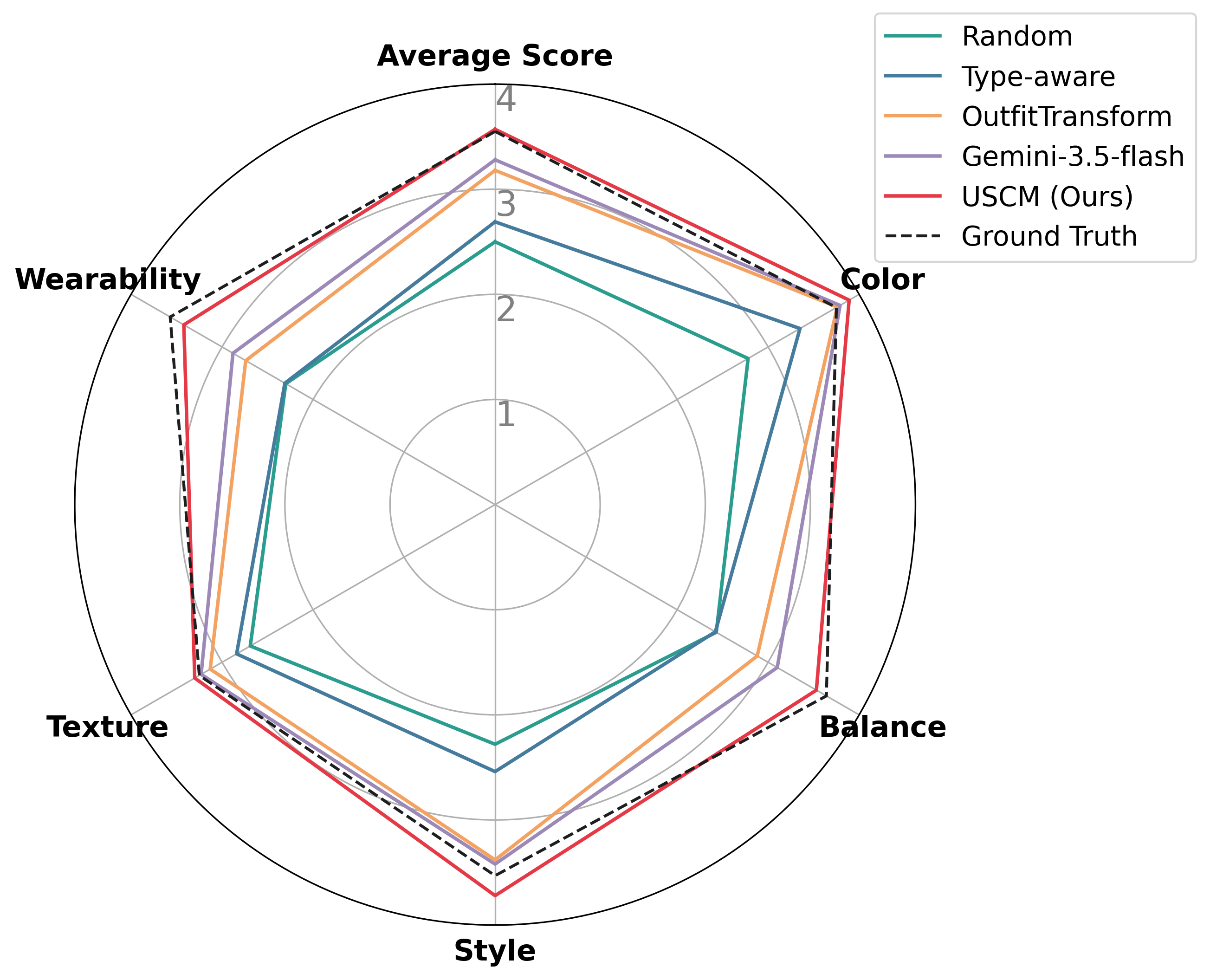}
  \vspace{-3mm}
  \caption{\textbf{Radar chart of VLLM Aesthetic Scores (VAS).} Composed outfits are evaluated across five aesthetic dimensions along with the overall average score.}
  \vspace{-4mm}
  \label{fig:radar_chart}
\end{wrapfigure}

\textbf{Baselines.}
We benchmark UAFM against four distinct paradigms encompassing random heuristics, greedy discriminative scorers, specialized transformers, and frontier vision-language models:
(1) \textbf{Random} arbitrarily selects candidates from the library to establish an empirical performance lower bound; 
(2) \textbf{Type-Aware} lacks a policy prior and greedily ranks $20$ randomly sampled items at each step using its discriminative scorer; 
(3) \textbf{OutfitTransformer} employs two separate models to handle compatibility modeling and next-item prediction respectively, where outfit compatibility is trained via a classification objective, and integrates them with beam search for iterative outfit composition;
and (4) \textbf{Gemini-3.5-flash}, an frontier vision-language model, directly interprets seed visual prompts and synthesizes targeted text queries to retrieve complementary items from the library.

\textbf{Results and Analysis:}
As presented in Table~\ref{tab:generation_quality}, UAFM establishes a clear state-of-the-art in FOC, outperforming all learning-based baseline models across all metrics. Specifically, UAFM achieves the highest Human Preference win rate ($46.17\%$), VLLM Aesthetic Score ($3.57$), and Cate. Align ($72.18\%$), while significantly reducing Category Duplication ($5.69\%$). UAFM also achieves the lowest Visual Redundancy ($1.20\%$) and fastest inference Latency ($0.13\text{~s}$). 
Furthermore, as shown in the radar chart in Figure~\ref{fig:radar_chart}, UAFM's fine-grained aesthetic profile closely aligns with GT outfits. UAFM surpasses GT outfits in the \textit{Color} and \textit{Style} dimensions, while slightly trailing them in \textit{Wearability} and \textit{Balance}. This discrepancy is quantitatively aligned with results of structural metrics: UAFM exhibits slightly higher Category Duplication ($5.69\%$ vs. $2.01\%$) and Visual Redundancy ($1.20\%$ vs. $0.15\%$) than GT outfits. 

As expected, the \textit{Random} strategy yields a low Visual Redundancy ($0.18\%$) and Latency ($0.0078\text{~s}$), but severely degrades outfit quality ($\text{VAS} = 2.50$) and Category Alignment ($42.34\%$) due to its unconstrained random nature. Another baseline \textit{Type-Aware} relies solely on a discriminative scoring mechanism, yielding poor outfit quality ($\text{VAS} = 2.69$) comparable to \textit{Random}. It shows that an explicit policy head is necessary for FOC.
\textit{OutfitTransformer} consistently lags behind UAFM across all quantitative dimensions. This performance gap stems primarily from its training objective: by optimizing compatibility via static supervised classification strictly on complete ensembles, it fails to evaluate the aesthetic potential of intermediate partial states, thereby suffering from a severe train-test mismatch during sequential composition.
\textit{Gemini-3.5-flash:} As a strong closed-source baseline, this model leverages extensive world knowledge to synthesize aesthetically appealing ensembles ($\text{VAS} = 3.28$). However, it frequently generates redundant categories, as evidenced by its high Category Duplication ($15.48\%$) and Visual Redundancy ($8.60\%$). Qualitative results, as shown in Figure~\ref{fig:qualitative_comparison}, reveal that this method over-attends to the seed item's attributes, repeatedly retrieving redundant pieces that compromise practical wearability ($2.88$ vs. Ours $3.14$). Furthermore, its massive scale incurs a substantially higher latency of $6.35\text{s}$ per outfit, which may limit real-time deployment, whereas UAFM requires a mere $0.13\text{s}$.

\begin{table}[t]
\centering
\caption{Performance comparison on downstream discriminative tasks on disjoint version of the Polyvore-Outfits dataset. Best results are highlighted in \textbf{bold}.}
\label{tab:baseline_comparison}
\setlength{\tabcolsep}{4pt}
\begin{tabular}{lccccc}
\toprule
Method & AUC & FITB (\%) & R@10 (\%) & R@30 (\%) & R@50 (\%) \\
\midrule
Type-Aware & 0.84 & 55.65 & 3.66 & 8.26 & 11.98 \\
SCE-Net & - & 53.67 & 4.41 & 9.85 & 13.87 \\
CSA-Net & 0.87 & 59.26 & 5.93 & 12.31 & 17.85 \\
LLM-DPO & 0.81 & 61.00 & - & - & - \\
OutfitTransformer & 0.90 & 69.01 & 10.51 & 20.33 & 27.64 \\
\midrule
\textbf{UAFM (Ours)} & \textbf{0.93} & \textbf{70.84} & \textbf{12.05} & \textbf{22.63} & \textbf{29.50} \\
\bottomrule
\end{tabular}
\end{table}

\subsection{Evaluation on Downstream Discriminative Tasks}
\label{subsec:sota_comparison}
To verify that UAFM captures robust aesthetic representations beyond sequential composition, we extend our evaluation to two conventional discriminative fashion tasks. Firstly, \textit{Fashion Compatibility} (FC) evaluation assesses the model's capacity to discriminate positive and negative ensembles, quantified by the \textit{Area Under the Curve (AUC)} metric. 
Secondly, \textit{Complementary Item} (CI) retrieval is assessed through two protocols: \textit{Fill-In-The-Blank (FITB)}, which tests the model's ability to select the most compatible item among four candidates to complete a partial outfit; and \textit{Recall@K (R@K)}, which measures the rank of the ground-truth target item against 3,000 distractors given a query outfit.
We compare UAFM with five state-of-the-art baselines:
\textit{Type-Aware}~\citep{vasileva2018learning} utilizes conditional embeddings to project items into category-specific similarity spaces, relying heavily on pairwise distance metric learning.
\textit{SCE-Net}~\citep{tan2019learning} learns visual similarity conditions dynamically without requiring explicit attribute supervision, though it remains restricted to pairwise feature comparisons.
\textit{CSA-Net}~\citep{lin2020fashion} proposes a category-based subspace attention network coupled with a set-wise outfit ranking loss to capture broader structural relationships.
\textit{LLM-DPO}~\citep{forouzandehmehr2024decoding} fine-tunes an LLM using Direct Preference Optimization to align generated recommendations with human aesthetic preferences.
\textit{OutfitTransformer}~\citep{sarkar2023outfittransformer} employs a self-attention mechanism with distinct task-specific tokens to learn global outfit representations under a traditional supervised classification paradigm.

As reported in Table~\ref{tab:baseline_comparison}, UAFM achieves consistent performance gains across all metrics on three tasks. 
In the FC task, our model reaches an AUC of $0.93$, outperforming baselines \textit{OutfitTransformer} ($0.90$) and \textit{CSA-Net} ($0.87$). 
A similar trend is observed in the FITB and retrieval tasks: UAFM attains a $70.84\%$ accuracy on FITB and $12.05\%$ R@10 in large-scale retrieval, compared to $69.01\%$ and $10.51\%$ for \textit{OutfitTransformer}, with consistent performance margins across R@30 and R@50 metrics.
We attribute these empirical results to the sequential MDP formulation. Traditional baselines model fashion compatibility as a static classification task, supervising the network exclusively on complete ensembles via a terminal binary signal. In contrast, our MDP formulation decomposes each outfit into a step-by-step trajectory, exposing the network to diverse partial contexts during training. This enables UAFM to receive intermediate supervision that quantifies the latent aesthetic potential of incomplete ensembles.

\subsection{Ablation Study}
\label{sec: ablation}

\subsubsection{Ablation on Training Objectives}
\label{ablation:train_obj}

To isolate the contribution of our step-wise value propagation mechanism in autonomous outfit composition, we conduct a controlled ablation study on the training objectives. While keeping the network backbone, policy head, and overall training recipe identical, we compare our post-decision state TD objective against a baseline variant where the value head is trained via a standard Binary Cross-Entropy (BCE) classification objective. This BCE variant treats compatibility estimation as a static binary classification task, supervising the value head strictly on complete outfits. Below, we demonstrate the superiority of our TD-based paradigm in terms of both dynamic termination accuracy and value landscape smoothness.

\begin{figure}[t]
    \centering
    \includegraphics[width=\linewidth]{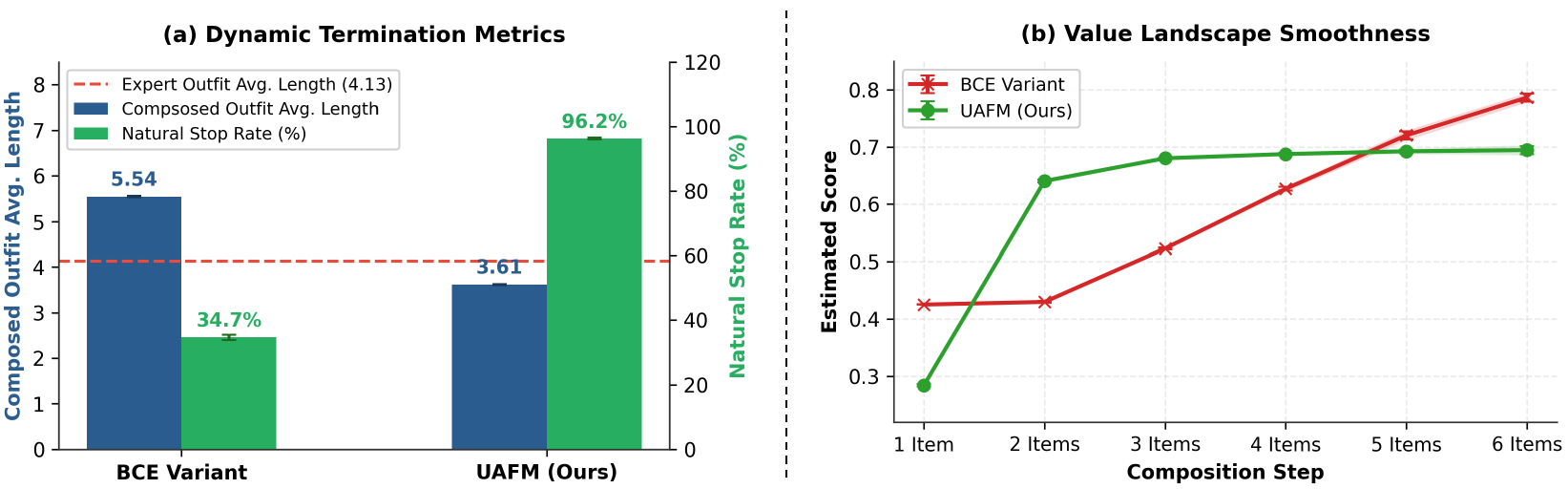}
    \vspace{-8mm}
    \caption{Ablation on training objectives.
    Part (a) compares average composed outfit length (blue bars) against GT outfit length (red dashed line) and natural stop rate (green bars). 
    Part (b) evaluate predicted aesthetic scores across composition steps with 3-fold cross-validation error bars.
    }
    \vspace{-5mm}
    \label{fig:ablation_on_training_method}
\end{figure}

\noindent\textbf{Experiment 1: Dynamic Termination.} This experiment evaluates the model's ability to autonomously halt composition. As items are sequentially added along each GT outfit trajectory in test set, composition dynamically terminates at step $t$ if the predicted marginal aesthetic gain drops below $0.0$. We quantify this capability using the \textit{natural stop rate}, defined as the proportion of trajectories that terminate at or before reaching their GT length. As shown in Figure~\ref{fig:ablation_on_training_method} (a), the BCE loss variant struggles with dynamic stopping, achieving a natural stop rate of only $34.7\%$. Because it cannot accurately evaluate intermediate states, it greedily over-composes items, resulting in an average outfit length of $5.54$—significantly overshooting the ground-truth average ($4.13$). Conversely, UAFM trained with TD loss achieves a highly reliable natural stop rate of $96.2\%$. Its average generated length of $3.61$ is much closer to the ground-truth length, demonstrating that the model effectively recognizes aesthetic saturation to prevent redundant additions.

\noindent\textbf{Experiment 2: Aesthetic Value Landscape.} Figure~\ref{fig:ablation_on_training_method} (b) reveals the underlying cause of the dynamic termination performance gap. Here, we evaluate how the aesthetic scores—predicted by the two value heads trained with different objectives—evolve as human-curated outfits are incrementally assembled from a single item to completion.
The BCE variant produces a largely linear, monotonically increasing landscape from the second step onward. Without the temporal awareness provided by value propagation, the BCE objective forces the network to degenerate into a naive feature accumulator. It erroneously equates the accumulation of items with a higher probability of matching the complete ground-truth label. Consequently, the network outputs a constant, positive marginal gain for every added item, failing to recognize aesthetic saturation and preventing the stopping threshold from ever triggering. 
In contrast, value head trained with the proposed post-decision TD objective can estimate the expected return of the composition trajectory. Its value landscape exhibits a steep initial gradient—correctly capturing the critical functional transition from a single item to a structurally complete core outfit. Once this core stylistic intent is fulfilled, the estimated value naturally plateaus, reflecting the principle of diminishing marginal aesthetic returns. This structural foresight directly enables UAFM to accurately trigger dynamic termination.

\begin{table}[t]
\centering
\caption{Performance comparison among the \textbf{BCE} variant, the \textbf{Separate} variant, and \textbf{UAFM}. All models share the identical backbone and training recipe, operating under dynamic termination without explicit length constraints. Best results are highlighted in \textbf{bold}.}
\label{tab:ablation_model_arch}
\small
\setlength{\tabcolsep}{2pt}
\begin{tabular}{lccccccccc}
\toprule
\multirow{2}{*}{\textbf{Architecture}} & \multicolumn{4}{c}{\textbf{Composed Outfit Quality}} & \multicolumn{5}{c}{\textbf{Discriminative Performance}} \\
\cmidrule(lr){2-5} \cmidrule(l){6-10}
 & VAS & Cate. Align \%  & Cate. Dup. \%$\downarrow$ & Vis. Red. \%$\downarrow$ & AUC & FITB& R@10  & R@30  & R@50\\
\midrule
BCE & 3.33 ± 0.02  & 55.23 ± 0.27  &  15.89 ± 0.11  & 3.70 ± 0.07 & \textbf{0.96} & 70.41 & 11.77 & 20.10 & 29.03 \\
Separate  &                3.27 ± 0.02          &  67.73 ± 0.29    &  18.01 ± 0.32 & 6.75 ± 0.22  & 0.84 & 68.57 & 10.09 & 19.72 & 26.25 \\
\textbf{UAFM}  & \textbf{3.60 ± 0.02} & \textbf{78.20 ± 0.49}    &  \textbf{1.29 ± 0.07} & \textbf{0.34 ± 0.04}  & 0.93 & \textbf{70.84} & \textbf{12.05} & \textbf{22.63} & \textbf{29.50}\\
\bottomrule
\end{tabular}
\end{table}

\subsubsection{Ablation on Architectures}
To evaluate the necessity of our unified architecture and the TD training objective for outfit composition quality, we quantitatively compare UAFM against two structural variants under dynamic termination settings. The \textit{BCE variant} trains the value head using conventional binary cross-entropy loss strictly on complete outfits. The \textit{Separate variant} utilizes the TD objective but trains the policy and value networks independently without parameter sharing in the backbone.
As shown in Table~\ref{tab:ablation_model_arch}, the \textit{BCE variant} achieves a high static AUC of $0.96$ but exhibits reduced performance on FITB and R@10 metrics. This indicates that supervising the value head strictly on complete ensembles leads to overfitting on terminal compatibility evaluation, which impairs its complementary item prediction capability. Furthermore, static binary classification induces a severe train-test mismatch during sequential composition, as the network fails to evaluate the aesthetic potential of intermediate partial states. Consequently, the BCE variant suffers from substantial category duplication ($15.89\%$ vs. $1.29\%$ for UAFM) and visual redundancy ($3.70\%$), noticeably degrading both structural category alignment ($55.23\%$) and overall aesthetic quality ($\text{VAS} = 3.33$).
The \textit{separate} variant's decoupled architecture leads to lower performance across metrics. In FOC, its VAS score decreases from $3.60$ to $3.27$, and Category Alignment drops to $67.73\%$. On downstream tasks, its AUC decreases to $0.84$. This decline can be attributed to representation misalignment between the unshared latent spaces: without a joint backbone, the independent policy head proposes candidate items that do not necessarily align with the value head's estimated high-reward regions, resulting in less cohesive compositions.

\section{Conclusion}
\label{sec:conclusion}

In this paper, we introduced the Unified Aesthetic Foresight Model (UAFM), a novel paradigm that addresses Fashion Outfit Composition as a deterministic Markov Decision Process. By coupling policy and value heads onto a shared encoder backbone, UAFM bridges the gap between outfit aesthetic estimation and complementary item retrieval. Crucially, we proposed a post-decision state TD objective, which enables the model to backpropagate sparse aesthetic signals to early intermediate states. This endows UAFM with aesthetic foresight, allowing it to evaluate partial outfits and autonomously execute dynamic termination upon reaching aesthetic saturation, thereby eliminating the reliance on rigid length constraints. Extensive experiments demonstrate that UAFM achieves state-of-the-art performance across both FOC and downstream discriminative fashion tasks. 
Despite its strong performance, our framework has several limitations. First, our framework lacks the integration of external contexts such as personalized user preferences and specific occasions. Second, trained on offline datasets like Polyvore, UAFM inherently captures static, average human curation preferences, lacking a mechanism to adapt to dynamically evolving fashion trends. Third, UAFM's robustness on noisy, in-the-wild user closet photos with complex backgrounds and lighting remains unverified. Fourth, the precise impact of the step penalty of $-0.1$ and terminal reward assignments across different outfit termination states on system dynamics remains to be thoroughly investigated. Future work will focus on incorporating contextual prompts, adapting to temporal trends, handling unconstrained visual inputs, and analyzing terminal reward sensitivity for more robust deployment.

\bibliography{iclr2027_conference}
\bibliographystyle{iclr2027_conference}

\newpage
\appendix
\section{Related Work}
\noindent\textbf{Modeling Fashion Compatibility.} 
Early research in fashion recommendation primarily focused on modeling the compatibility score of a complete outfit. Initial approaches utilized pairwise metric learning~\citep{han2017learning} and category-aware subspace projections~\citep{vasileva2018learning} to map complementary items closely in a latent space. To capture higher-order structural relationships within an outfit, subsequent works introduced graph neural networks~\citep{cui2019dressing, saed2025hybrid} and hierarchical relation networks~\citep{pang2025towards}. Recently, large multimodal models have also been adapted to evaluate outfit harmony~\citep{chang2025using}. However, these methods inherently function as static, passive evaluators of fixed ensembles, lacking the active policy required to retrieve and compose complementary items for FOC. Moreover, since they are mainly optimized via supervised classification learning paradigms on complete outfits, they suffer from a train-test mismatch when applied to evaluate intermediate partial outfits.

\noindent\textbf{Retrieval-based and Generative Outfit Composition.} 
To move beyond passive scoring, generative methods have been proposed to actively retrieve or synthesize outfits. Retrieval-based frameworks, such as SCE-Net~\citep{tan2019learning} and OutfitTransformer~\citep{sarkar2023outfittransformer}, attempt to predict latent embeddings to index complementary items from a library. Recent advancements also include LLM-driven agents~\citep{forouzandehmehr2024decoding} and image-to-image translation models for generating retrieval templates~\citep{attimonelli2024fashion}. Alternatively, direct synthesis methods leverage diffusion models and GANs to generate pixel-level outfit images~\citep{zhou2022learning, xu2024diffusion, yu2025fashiondpo}. 
However, direct synthesis methods require explicit structural priors (e.g., reference masks or category prompts) and lack the capability to navigate a library to compose outfits autonomously. Moreover, single-shot set generation paradigms suffer from a severe \textbf{grounding mismatch} when projected onto a finite, discrete garment library. By simultaneously predicting multiple target embeddings in an unconstrained latent space, independent post-hoc item retrieval induces unmonitored quantization errors, causing the retrieved physical items to mutually clash.
And for retrieval-based methods, a critical architectural bottleneck remains: they typically treat next-item retrieval and overall compatibility evaluation as disjoint tasks optimized via decoupled models. This isolation induces conflicting latent spaces, where the retrieval policy may propose items that the compatibility scorer penalizes.

\noindent\textbf{Reinforcement Learning in Fashion and Recommendation.} 
While Reinforcement Learning (RL) and Markov Decision Processes (MDPs) have been broadly applied in recommendation systems, these paradigms uniquely model the \textit{external user's behavior}—where the state represents user interaction history, and the environment yields rewards based on stochastic user feedback (e.g., click-through rates). In contrast, within the fashion domain, MDP formulations have been largely restricted to operational decision-making such as optimizing dynamic pricing for seasonal products~\citep{aviv2005partially}, modeling cognitive human choice sets~\citep{li2012cognitive}, and managing attribute-level inventory risk~\citep{koren2026dynamic}. These prior works operate over structured, low-dimensional state and action spaces, where decisions correspond to finite operational adjustments. 
To the best of our knowledge, our work is the first to formalize the \textit{intrinsic visual composition} of an unordered fashion ensemble as a fully-observed, deterministic MDP. Rather than modeling external user choices, our environment is defined by the outfit state itself. By introducing the post-decision state TD objective, UAFM alleviates the sparse reward issue in outfit composition, equipping the model with aesthetic foresight and enabling dynamic termination based on marginal aesthetic gains.

\section{Theoretical Analysis of Post-Decision State TD Objective}
\label{apx:theoretical_analysis}
Under conventional terminal-only supervision, intermediate states $S_t^a$ receive zero gradients from the sparse terminal reward $y \in \{+1.0, -1.0\}$, inducing a severe train-test mismatch during sequential composition process. We theoretically demonstrate how our Post-Decision State TD objective resolves this by analyzing its gradient dynamics. The expected TD loss is defined as:
\begin{equation}
    \mathcal{L}_{\text{TD}}(\mathbf{w}) = \frac{1}{2} \mathbb{E}_{(S_t^a, S_{t+1}^a, R_{t+1}) \sim \mathcal{D}} \left[ \left(\hat{v}(S_t^a; \mathbf{w}) - Y_t\right)^2 \right],
\end{equation}
where $\mathbf{w}$ denotes the trainable parameters of the shared Transformer backbone and value head. The parameter update is driven by the gradient:
\begin{equation}
    \nabla_\mathbf{w} \mathcal{L}_{\text{TD}}(\mathbf{w}) = - \mathbb{E}_{\mathcal{D}} \left[ \delta_t \nabla_\mathbf{w} \hat{v}(S_t^a; \mathbf{w}) \right],
\end{equation}
where $\delta_t = Y_t - \hat{v}(S_t^a; \mathbf{w})$ is the TD error.

At the terminal step ($t = T-1$), the target is the deterministic global aesthetic score $Y_{T-1} = y$. The gradient explicitly anchors the complete outfit's value:
\begin{equation}
    \nabla_\mathbf{w} \mathcal{L}_{\text{TD}}^{(T-1)} = - \underbrace{\left(y - \hat{v}(S_{T-1}^a; \mathbf{w})\right)}_{\text{Terminal error } \delta_{T-1}} \nabla_\mathbf{w} \hat{v}(S_{T-1}^a; \mathbf{w}).
\end{equation}

For any intermediate state ($t < T-1$), the target is bootstrapped via the slowly-updating target network $\mathbf{w}^-$ as $Y_t = r + \gamma \hat{v}(S_{t+1}^a; \mathbf{w}^-)$. The gradient minimizes the discrepancy between the current state value and the discounted future potential:
\begin{equation}
    \nabla_\mathbf{w} \mathcal{L}_{\text{TD}}^{(t)} = - \underbrace{\left( r + \gamma \hat{v}(S_{t+1}^a; \mathbf{w}^-) - \hat{v}(S_t^a; \mathbf{w}) \right)}_{\text{Intermediate error } \delta_t} \nabla_\mathbf{w} \hat{v}(S_t^a; \mathbf{w}).
\end{equation}

Assuming optimal convergence, recursively expanding the value estimation reveals that the effective target $Y_t$ for an early intermediate state mathematically approximates its discounted cumulative return:
\begin{equation}
    Y_t \approx \left( \sum_{k=t}^{T-2} \gamma^{k-t} r \right) + \gamma^{T-1-t} y
\end{equation}

This closed-form formulation explicitly decouples the intermediate state value into two competing forces, yielding two fundamental advantages for the UAFM framework:

\textbf{1. Aesthetic Foresight and Dynamic Termination:} The formulation mathematically decomposes the intermediate value into \textit{Length Regularization} (a monotonically increasing penalty since $r < 0$, discouraging redundancy) and \textit{Aesthetic Bounding} (the exponentially decayed terminal reward $y$). By densely propagating $y$ backward ($y \to \hat{v}(S^a_{T-1}) \dots \to \hat{v}(S^a_1)$), this structural credit assignment grants the value head aesthetic foresight. Consequently, the model proceeds with the composition only if the projected aesthetic gain outweighs the cumulative step penalty. This tension mathematically guarantees that the value head can accurately estimate marginal utility, triggering early termination exactly when aesthetic saturation is reached.

\textbf{2. Synergistic Manifold Smoothing:} Because UAFM shares the Transformer backbone across the policy and value heads, the parameters $\mathbf{w}$ govern a joint latent representation manifold. Here, the TD error $\delta_t$ acts as a temporal regularizer. By enforcing Bellman consistency ($\hat{v}(S_t^a) \approx r + \gamma \hat{v}(S_{t+1}^a)$), the gradient updates restrict the latent representations of consecutive states to be contiguous and smoothly graded. This temporal smoothing intrinsically aligns the item retrieval intent with the compositional value trajectory, reducing semantic fragmentation and underlying the unified architecture's performance gains over decoupled variants.

\section{Implementation Details}
\label{apx:implementation}
\subsection{UAFM Architecture and Hyperparameters}
\textbf{Model Architecture.} 
For the item encoder, we adopt the frozen pre-trained FashionCLIP~\citep{Chia2022} to extract 512-dimensional normalized visual features for all garment images. To ensure practical deployability in real-world scenarios where textual metadata is often unavailable, we deliberately omit text feature extraction and rely exclusively on visual representations.
The shared Transformer encoder backbone consists of $6$ encoder layers, $16$ attention heads, a hidden dimension of $512$, and a feed-forward dimension of $2024$ with GELU activations.
To ensure a fair comparison on FOC and downstream discriminative tasks, the baseline \textit{OutfitTransformer} employed utilize the exact same frozen pre-trained FashionCLIP as item feature extractor.

\textbf{Offline Training Setup.} 
We train UAFM for $50,000$ steps with a batch size of $1,000$ on the Polyvore-Outfits dataset. Each batch contains a balanced $1:1$ ratio of positive and negative transition pairs. 
For offline trajectory decomposition, an unordered outfit ensemble is randomly permuted into sequential trajectories during training to ensure permutation robustness.
We optimize the model using AdamW with a peak learning rate of $2 \times 10^{-5}$, and weight decay of $0.01$. 
We adopt the OneCycleLR scheduler~\citep{al2022scheduling} with a cosine annealing strategy, where the warm-up phase accounts for $30\%$ of total steps. 
In the post-decision TD objective, the discount factor is set to $\gamma = 0.9$, and the target network $\mathbf{w}^-$ is updated every $100$ steps via Polyak averaging~\citep{polyak1992acceleration} with $\tau_{\text{polyak}} = 0.005$. 
In the InfoNCE loss, the temperature parameter is set to $\tau_{\text{nce}} = 0.2$, and the joint loss balance factor is $\lambda = 1.0$.
All training was conducted on a single NVIDIA GeForce RTX 4090 D GPU (24GB VRAM), taking approximately $10$ hours to complete.

\subsection{Negative Outfit Construction Strategy}
\label{apx:appendix_noc}
To enable the value head to effectively discriminate aesthetic dissonance and structural invalidity, we devise a negative outfit construction strategy for training. For each positive human-curated outfit of length $N$, a negative outfit ($y = -1.0$) of equal length is synthesized by randomly executing one of three distinct negative generators:

\begin{enumerate}
    \item \textbf{Intra-Category Item Substitution:} Each item in the positive outfit is replaced by a randomly sampled item from the exact same category. This preserves the functional category template (e.g., Top--Bottom--Shoes) while disrupting the fine-grained stylistic compatibility among individual items.
    
    \item \textbf{Visual Redundancy Collapsing:} An item is randomly selected as an anchor, and its $N-1$ nearest visual neighbors in CLIP feature space are retrieved via similarity search. This constructs visually homogeneous outfits suffering from excessive visual redundancy.
    
    \item \textbf{Category Redundancy Collapsing:} An item is randomly chosen as an anchor, and the remaining $N-1$ slots are populated entirely with distinct items drawn from the anchor's category. This yields structurally invalid outfits composed of duplicate functional types (e.g., an outfit consisting entirely of tops).
\end{enumerate}


\subsection{Inference and Beam Search Mechanics}
During autonomous outfit composition, UAFM is coupled with a beam search algorithm with candidate retrieval size $K = 10$ and beam width $B = 10$. 
At each step $t$, for each trajectory in the beam, the Policy Head generates the query representation of the target item, from which $K$ nearest neighbors are retrieved from the library based on latent feature similarity.
The Value Head then evaluates the post-decision state value $\hat{v}(S_{t+1}^a)$ for each candidate-augmented ensemble $S_{t+1}^a = S_t \cup \{a\}$. Trajectories in the beam are subsequently ranked and filtered strictly based on their predicted post-decision values $\hat{v}(S_{t+1}^a)$
The composition process dynamically terminates when the maximum marginal value gain falls below the threshold $\tau_{\text{stop}} = 0.0$.
Note that except for the main FOC evaluation in Table~\ref{tab:generation_quality}, where the generation length is constrained to match ground-truth length for fair baseline comparison, no length restrictions or category-filtering rules are imposed across all other experiments.

\section{Ablation Study on Impact of Decoding Strategies}
\label{sec:appendix_search_strategy}

To explicitly investigate how heuristic decoding algorithms influence the composition process and to verify whether the generative capability of the policy head renders the value head redundant, we conduct an ablation study comparing four decoding strategies utilizing the UAFM framework. Specifically, \textbf{Value-Greedy} performs local optimization by retrieving top-10 candidates at each step via the policy head and greedily selecting the single highest-scoring afterstate evaluated by the value head. \textbf{Beam Search} maintains $10$ parallel trajectories, expanding each beam by 10 candidates and using the value head to prune the combined set back to the top-$10$ states. \textbf{Reranking} adopts a compose-then-rerank scheme: it stochastically samples $10$ candidate trajectories by drawing next-item candidates from the policy head's similarity distribution (scaled by temperature $\tau=0.2$). During autonomous composition, each candidate trajectory dynamically terminates when the value head's predicted marginal value gain falls below a threshold $0.0$. Finally, the value head re-evaluates all $M=10$ dynamically generated ensembles in batch to select the single highest-scoring outfit. Finally, to isolate the functional contribution of the value head, we introduce a \textbf{Policy-Only} baseline. This method bypasses the value head entirely by querying the policy head and deterministically appending the first retrieved novel candidate to the partial outfit at each step. Because it lacks the value head's dynamic termination capability based on marginal aesthetic utility, the Policy-Only method is explicitly constrained to compose items until reaching the ground-truth outfit length to ensure a fair structural comparison. 

\begin{table}[t]
\centering
\caption{Comparison of different decoding strategies utilizing UAFM for outfit composition.}
\label{tab:search_ablation}
\small
\setlength{\tabcolsep}{2pt}
\begin{tabular}{l | c|c|c|c|c|c|c|c|c|c}
\toprule
\multirow{2}{*}{Strategy}  & \multirow{2}{*}{Latency}  & \multirow{2}{*}{Cate. Align \%$\uparrow$} & \multirow{2}{*}{Cate. Dup. \%$\downarrow$} & \multirow{2}{*}{Vis. Red. \%$\downarrow$} & \multicolumn{6}{c}{\textbf{VLLM Aesthetic Score $\uparrow$}}\\
\cmidrule(l){6-11}
& & & & & Col. & Bal. & Sty. & Tex. & Wear. & Overall\\
\midrule
Value-Greedy  & 0.028s & \textbf{78.99} ± 0.48  & 1.01 ± 0.07 & 0.37 ± 0.02  & 3.93 & 3.47 & 3.77 & 3.24 & \textbf{3.52}  & 3.59 ± 0.02 \\
Reranking  & 0.24s & 78.80 ± 0.43  & 1.07 ± 0.06 & 0.39 ± 0.06 & 3.93 & 3.48 & \textbf{3.78} & \textbf{3.31} & 3.48 & \textbf{3.60} ± 0.03 \\
Beam Search & 0.13s & 78.20 ± 0.49 & \textbf{1.29} ± 0.07 & \textbf{0.34} ± 0.04 & \textbf{3.94} & \textbf{3.50} & 3.77 & 3.30 & 3.50 & \textbf{3.60} ± 0.03\\
Policy-Only & \textbf{0.021}s & 53.86 ± 0.14 & 6.82 ± 0.21 & 2.78 ± 0.13 & 3.91 & 3.26 &3.68&3.26& 3.03 &3.43 ± 0.03\\
\bottomrule
\end{tabular}
\end{table}

Table~\ref{tab:search_ablation} demonstrate two primary findings regarding the UAFM architecture and the choice of decoding strategies. First, the \textit{Policy-Only} strategy exhibits a notable performance degradation across all structural and aesthetic metrics. Without the value head to evaluate holistic ensemble compatibility and provide aesthetic foresight, Cate. Align drops significantly to 53.86\%, while categorical duplication and visual redundancy increase to 6.82\% and 2.78\%, respectively. Consequently, yielding the lowest overall VLLM Aesthetic Score (VAS) of 3.43. This confirms that relying solely on the policy head is insufficient for cohesive outfit composition, thereby validating the necessity of the value head for maintaining structural rules and overall aesthetic harmony. Second, when the value head is integrated, the three distinct decoding strategies (\textit{Value-Greedy}, \textit{Reranking}, and \textit{Beam Search}) achieve highly similar generative performances. All three methods maintain comparable VAS and structural metrics, with Category Align remaining tightly bounded between 78.20\% and 78.99\%. This consistency indicates that the final composition quality is fundamentally driven by the robust multi-task representation and accurate aesthetic value estimations of the UAFM backbone, rather than the specific complexity of the underlying decoding strategies.

\section{Baseline Configurations and Evaluation Protocols via VLLM}
\subsection{VLLM Baseline Configuration and Prompt Design}
\label{apx:gemini_foc}

To construct a representative Vision Large Language Model (VLLM) baseline for Fashion Outfit Composition, we integrate \texttt{gemini-3.5-flash} into a hybrid text-guided retrieval framework. Given a partial outfit $S_{\text{init}}$ and a target outfit length $L_{\text{target}}$, the end-to-end execution proceeds in three streamlined steps:
\textbf{(1)~Multimodal Context Assembly:} product images in $S_{\text{init}}$ are composite-stitched into a single square visual prompt alongside their concatenated raw text descriptions $D_{\text{init}}$.
\textbf{(2)~Structured Description Generation:} \texttt{gemini-3.5-flash} synthesizes detailed visual descriptions for $N_{\text{fill}} = L_{\text{target}} - |S_{\text{init}}|$ missing complementary items under strict JSON schema enforcement.
\textbf{(3)~Text-Guided Candidate Retrieval:} the generated descriptions are mapped into a continuous latent space via Fashion-CLIP to query candidate items from the library using $k$-NN search ($k=5$) with dynamic deduplication.

\paragraph{System Instruction and Generation Setup.}
We configure \texttt{gemini-3.5-flash} with structured output enforcement (\texttt{response\_mime\_type="application/json"}) at a fixed temperature of $\tau = 0.7$. The system instruction and user prompt template are defined as follows:

\begin{quote}
\small
\textbf{System Instruction:} \\
\texttt{"You are a professional fashion stylist. Based on the images provided by the user, please recommend exact \{num\_to\_fill\} items that would enhance the overall harmony of the outfit. Recommended items MUST come from these valid categories: \{valid\_categories\}. Your answer must be in English. Start item's description is \{item\_description\}"}
\end{quote}

\paragraph{Structured Output Schema.}
To guarantee deterministic programmatic parsing for downstream vector encoding, the API response is strictly constrained to the following structured JSON format:

\begin{quote}
\small
\begin{verbatim}
{
  "recommendations": [
    {
      "description": "A detailed visual description of the recommended 
                      item, including its category, color, material, 
                      and stylistic features to facilitate retrieval."
    }
  ]
}
\end{verbatim}
\end{quote}

\subsection{VLLLM Aesthetic Evaluator Setup}
\label{apx:vlm_evaluator}

To evaluate the visual aesthetic quality of composed outfits, we employ \texttt{gemini-3.7-flash} as an automated VLLM judge. Crucially, our evaluation framework is \textbf{anchored on the seed item}, directly reflecting real-world styling constraints where complementary garments must harmonize with it rather than merely among themselves.

We enforce structured JSON generation using Pydantic schema validation at a low decoding temperature ($\tau = 0.2$) to eliminate parsing errors and minimize evaluation variance. The output schema strictly defines five fine-grained 1--5 Likert-scale aesthetic dimensions alongside a qualitative justification. The system instruction and user prompt template are defined as follows:

\begin{quote}
\small
\textbf{System Instruction:} \\
\texttt{"You are an expert fashion stylist acting as an objective judge for composed outfits. You will be provided with a composite image of an outfit and its textual metadata. CRITICAL RULE: The first item (Item 1, top-left in the image) is the FIXED SEED ITEM. Your evaluation MUST anchor entirely on this Seed Item. If the subsequently added items (Item 2, 3, 4, etc.) have a great style together but CLASH with Item 1, the outfit is a FAILURE and must be scored poorly.\\
\\
Five assessed dimensions are defined as follows:\\
- Color Harmony: color coordination, contrast, and tonal synergy relative to the seed item.\\
- Balance \& Proportion: silhouette, cut, and visual volume hierarchy.\\
- Style Cohesion: aesthetic consistency, heavily penalizing items that clash with the seed item.\\
- Texture \& Detail: material compatibility and textural depth across garments.\\
- Wearability: real-world practicality, contextual appropriateness, and functional logic.\\
\\
Score each dimension from 1 to 5:\\
- 5 (Flawless): Added items perfectly elevate and match the Seed Item.\\
- 4 (Good): Highly cohesive with the Seed Item, minor imperfections.\\
- 3 (Acceptable): Average daily wear, cohesive but uninspired.\\
- 2 (Poor): Added items mismatch the Seed Item's style, color, or proportion.\\
- 1 (Disastrous): Severe conflict with the Seed Item (e.g., formal blazer with beach shorts)."}
\end{quote}

\begin{quote}
\small
\textbf{User Prompt Template:} \\
\texttt{"Evaluate the fashion outfit shown in the image based on its visual representation and metadata. REMEMBER: Item 1 is the FIXED SEED ITEM. Evaluate how well the rest of the outfit works WITH Item 1.\\
\\
--- Item Descriptions ---\\
\{description\}\\
-------------------------\\
\\
Provide scores from 1 to 5 for all 5 dimensions and brief reasoning."}
\end{quote}

\subsection{Human Preference Evaluation Protocol}
\label{apx:human_study}
In this section, we describe the acquisition methodology and data collection protocol for the Human Preference (H. Pref.) evaluations presented in Table~\ref{tab:generation_quality}. The detailed setup, questionnaire design, participant demographics, and statistical validation are elaborated below.

\begin{figure}[h]
    \centering
    \includegraphics[width=0.95\linewidth]{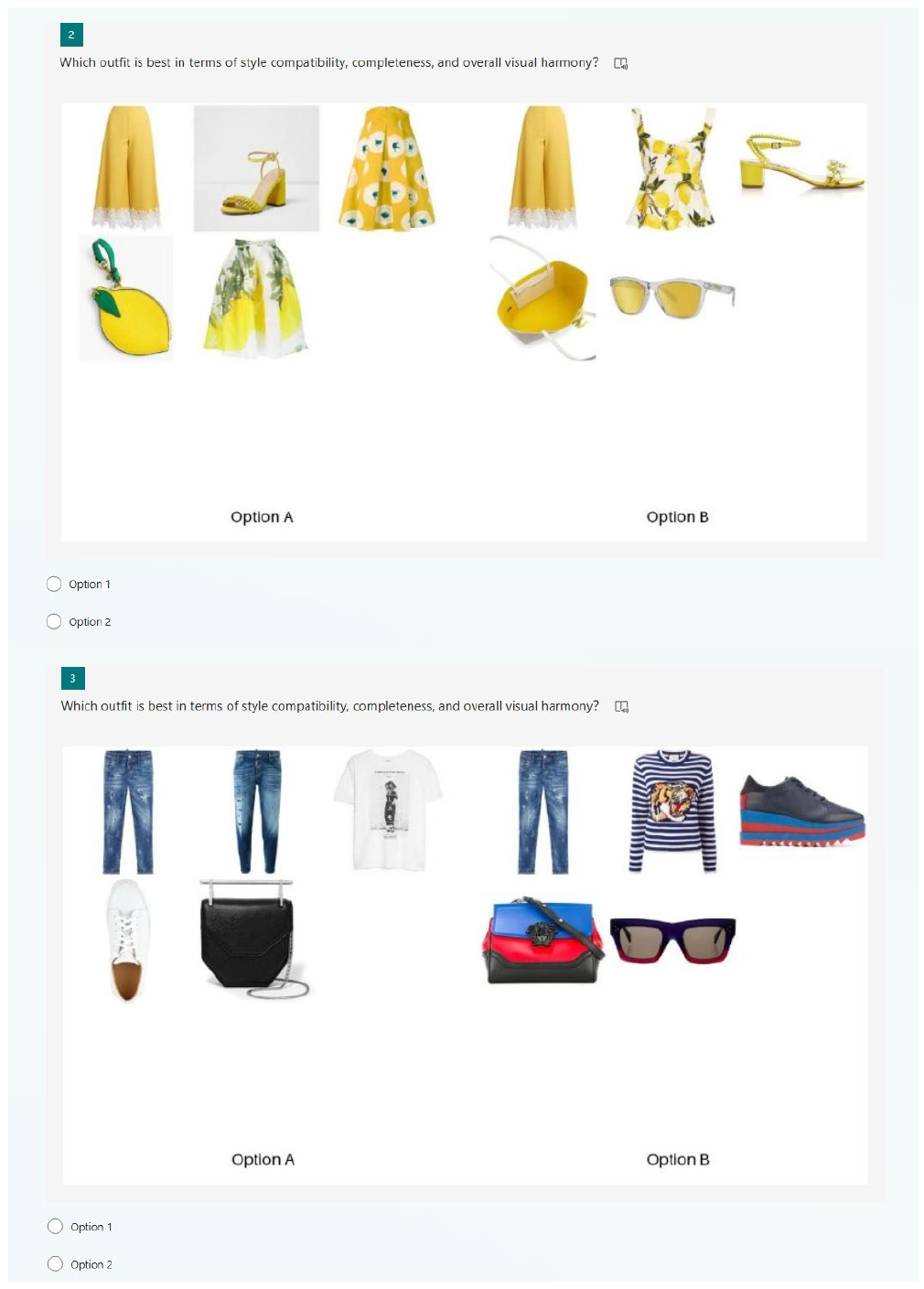}
    \caption{\textbf{Interface screenshot of the Human Preference A/B testing questionnaire on Microsoft Forms.} Evaluators conduct blind pairwise comparisons between outfits composed by different algorithms under standardized visual criteria.}
    \label{fig:human_study_interface}
\end{figure}

To compare UAFM with two competitive baselines (OutfitTransformer and Gemini-3.5-flash), we conducted a full all-pairs comparison across randomly sampled test cases. For each comparison trial, a seed item is randomly selected from the test set and provided identically to both methods to assemble their respective outfits, which are then presented side-by-side for comparative assessment. As shown in Figure~\ref{fig:human_study_interface}, the questionnaire was hosted on Microsoft Forms. For every question, candidates were asked to select the superior outfit based on the following explicit criterion:

\textit{"Which outfit is best in terms of style compatibility, completeness, and overall visual harmony?"}

To eliminate position bias and ensure a strict double-blind evaluation, the left/right presentation order of Option A and Option B was fully randomized and anonymized for all pairwise comparisons. Data was collected via a hybrid deployment model combining online link distribution for general users with on-site field surveys of experienced fashion store guides. In total, 40 unique evaluators participated, yielding 1,200 valid paired choices across 30 test comparisons.

The aggregated preference win rates reported in Table~\ref{tab:baseline_comparison} reflect the proportion of choices favoring each methods. To verify that the performance margins are statistically meaningful rather than artifacts of random variance, we performed a Chi-Square ($\chi^2$) goodness-of-fit test on the preference distributions. The statistical analysis confirms that UAFM significantly outperforms baseline paradigms ($\chi^2 = 89.92, df = 2, p < 0.001$), strongly validating the reliability and significance of our human preference evaluation.

\section{Additional Qualitative Results}
\label{apx:qualitative_results}
\begin{figure}[t]
    \centering
    \includegraphics[width=1.0\linewidth]{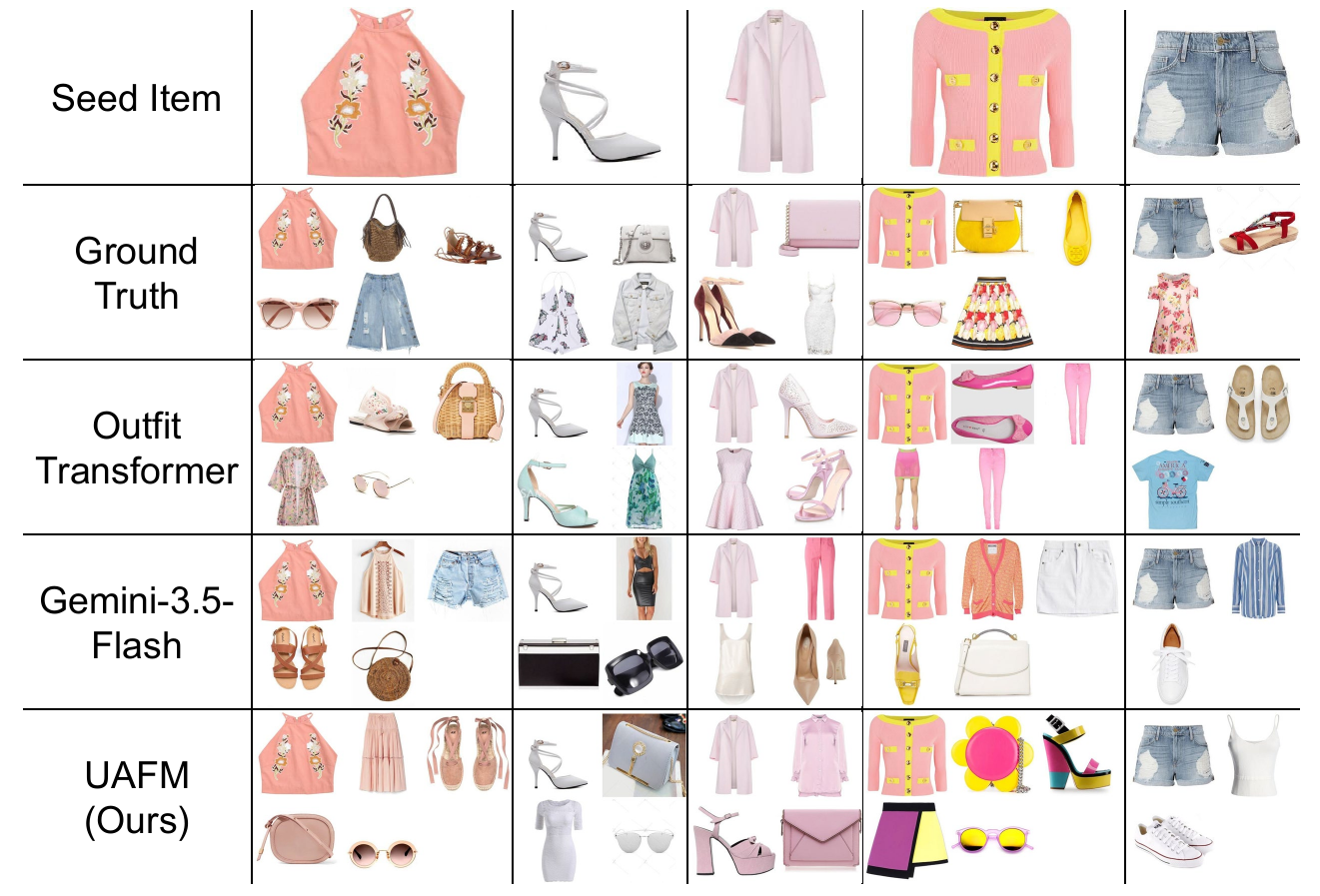}
    \caption{\textbf{Qualitative comparison of composed outfits across different paradigms.} Given an initial seed garment, each system sequentially retrieves complementary items to assemble a complete ensemble. Compared with baseline paradigms, UAFM (Ours) achieves superior stylistic cohesion, color harmony, and category diversity, closely resembling human compositions (\textit{Ground Truth}).}
    \label{fig:qualitative_comparison}

\end{figure}

Figure~\ref{fig:qualitative_comparison} presents qualitative comparison results of composed fashion outfits between our method and baseline paradigms given initial seed items.


\end{document}